\documentclass{article}

\PassOptionsToPackage{numbers}{natbib}
\usepackage[final]{neurips_data_2024}
\usepackage{wrapfig}
\usepackage{array}
\usepackage{hyperref}
\usepackage{pifont}
\usepackage{bm}
\usepackage{enumitem}
\newcommand{\cmark}{\ding{51}}
\newcommand{\xmark}{\ding{55}}

\usepackage[textsize=tiny]{todonotes}

\usepackage[utf8]{inputenc} 
\usepackage[T1]{fontenc}    
\usepackage{hyperref}       
\usepackage{url}            
\usepackage{booktabs}       
\usepackage{amsmath}
\usepackage{amsfonts}       
\usepackage{nicefrac}       
\usepackage{microtype}      
\usepackage{xcolor}         
\usepackage{changepage}
\usepackage{amsmath}
\usepackage{algorithm}
\usepackage{algpseudocode}
\usepackage{tikz}
\usepackage{multirow}

\title{\textsf{QGym}: Scalable Simulation and Benchmarking of Queuing Network Controllers}

\author{ Haozhe Chen$^{1,*}$ \quad Ang Li$^{1,*}$ \quad Ethan Che$^{2,}$\thanks{Equal contribution.} \\
\quad \textbf{Tianyi Peng}$^2$ \quad \textbf{Jing Dong}$^2$ \quad \textbf{Hongseok Namkoong}$^2$\\ \\
$^1$Columbia University \quad $^2$Columbia Business School\\
\texttt{\{haozhe.chen, al4263\}@columbia.edu}\\
\texttt{\{eche25, tianyi.peng, jing.dong, namkoong\}@gsb.columbia.edu}
}

\begin{document}

\maketitle
\

\begin{abstract}
Queuing network control determines the allocation of scarce resources to manage congestion, a fundamental problem in manufacturing, communications, and healthcare. Compared to standard RL problems, queueing problems are distinguished by unique challenges: i) a system operating in continuous time, ii) high stochasticity, and iii) long horizons over which the system can become unstable (exploding delays). To spur methodological progress tackling these challenges, we present an open-sourced queueing simulation framework, \textsf{QGym}, that benchmark queueing policies across realistic problem instances. Our modular framework allows the researchers to build on our initial instances, which provide a wide range of environments including parallel servers, criss-cross, tandem, and re-entrant networks, as well as a realistically calibrated hospital queuing system. 
\textsf{QGym} makes it easy to compare multiple policies, including both model-free RL methods and classical queuing policies. Our testbed complements the
traditional focus on evaluating algorithms based on mathematical guarantees in idealized settings,
and significantly expands the scope of empirical benchmarking in prior work.
\textsf{QGym} code is open-sourced at \href{https://github.com/namkoong-lab/QGym}{\texttt{https://github.com/namkoong-lab/QGym}}.
\end{abstract}


\section{Introduction}

Queuing network control is a fundamental control problem in managing congestion in job-processing systems, such as semiconductor manufacturing fabrication plants, communications networks, cloud computing facilities, call centers, healthcare delivery systems, ride-sharing platforms, and limit-order books \citep{shanthikumar2007queueing, harchol2013performance, neely2022stochastic, aksin2007modern,armony2015patient, banerjee2022pricing, cont2010stochastic}. In a typical queuing system, jobs arrive at random intervals, wait in queues until an available server can service them, and then either leave the system or move on to another queue for further processing. 
The stochastic workload is the defining challenge in queueing: the variability inherent in real-world systems makes the time it takes to process jobs random. 
To manage performance objectives such as minimizing processing delays, balancing workloads, and improving overall service quality and efficiency, a good controller must dynamically allocate resources accounting for future stochasticity. The ability to plan is especially crucial for systems that experience varying levels of congestion over time. 


Routing/scheduling control (i.e., matching jobs with servers) in industrial-scale systems is challenging due to several factors. First, queueing networks can be large and complex, with many different job classes and server types (see, e.g., \cite{harchol2013performance,armony2015patient}). 
The processing speeds can depend on the server and job types, which must be taken into account to make effective scheduling decisions.
Second, in systems such as semiconductor fabrication that involve multiple stages of \emph{sequential processing}, congestion can occur at various points in the queuing network, creating bottlenecks that slow down the entire system. Third, workloads are highly non-stationary, featuring predictable and unpredictable demand spikes over time.

\begin{figure}[t]
\centering
\vspace{-1cm}
  \hspace{-2.9cm}  
  
  \includegraphics[width=1.\textwidth]{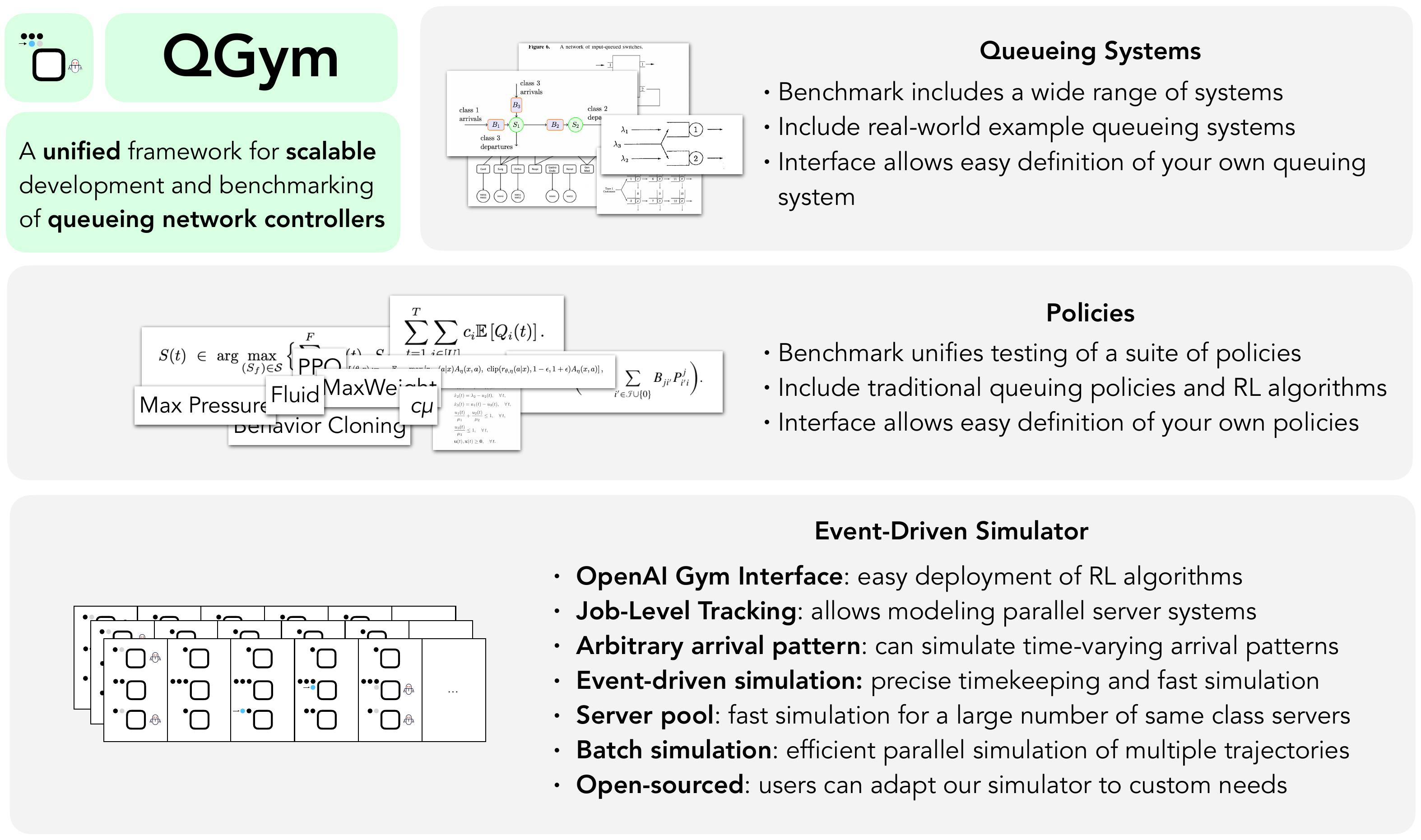}
  \caption{Highlights of \textsf{QGym} framework for developing and benchmarking queuing algorithms. \textsf{QGym} provides an event-driven simulator and benchmarks a wide range of queuing policies and systems. \textsf{QGym} interface also allows users to easily specify new queuing policies and systems.}
\end{figure}

There is a large body of work devoted to developing good routing/scheduling policies for queueing networks.
The traditional focus of methodological development has been on simple policies with good theoretical performance guarantees in specific network structures. These include (1) load-balancing rules such as joining the shortest queue, and variations such as the power-of-$d$ choices where one randomly samples $d$ queues and sends the job to the shortest one among them \citep{mitzenmacher2001power}; 
(2) scheduling rules to minimize delay cost such as shortest processing time first, and variations of it such as the $c\mu$-rule \citep{buyukkoc1985cmu} 
and the generalized $c\mu$-rule
  \citep{mandelbaum2004scheduling}
 when we do not know the exact job size and different job classes are associated with different delay costs; 
and (3) policies that achieve the maximum stability region such as MaxWeight \citep{stolyar2004maxweight} and maximum pressure policies \citep{dai2005maximum}. 

While easy to implement and interpret, these policies are restrictive to specific queuing architectures or objectives, not data-driven, and difficult to adapt to network-specific features and nonstationarity. As a result, there is a growing interest in using black-box reinforcement learning (RL) techniques to learn queuing controllers in a more data-driven manner, which can better handle realistic settings featuring complex networks with non-stationary workloads. 

However, queuing network control poses unique challenges, necessitating new methodological innovation over existing RL algorithms.
Compared to typical robotics or game-playing environments, queueing problems have high stochasticity and longer horizons over which the system can become unstable (exploding delays).
As a result, model-free RL algorithms have been observed to suffer from instability and substantial stochastic noise in these environments~\citep{singh1996reinforcement, liu2022rl, dai2022queueing}. 

Another unique challenge is that the queueing system naturally evolves in continuous time, in contrast to the standard discrete time formulation of RL problems. Existing studies assume inter-arrival times and job processing times are exponentially distributed~\citep{singh1996reinforcement, liu2022rl, dai2022queueing}. This so-called Markovian assumption is invoked to represent the queuing network
as a discrete-time Markov Decision Process (MDP) with queue lengths as state variables ~\citep{neely2010stochastic}. However, this assumption frequently does not hold in practice, as realistic event times typically exhibit higher variances~\citep{paxson1995wide}.


Prior works~\citep{singh1996reinforcement, liu2022rl, dai2022queueing} demonstrate the performance of RL algorithms on  a small number of problem instances and there is a lack of a common simulation environment or benchmark suite that provides comprehensive evaluations of baselines, including RL algorithms and theory-driven queuing policies. 
To address this need, we develop a flexible queuing simulation framework (open-sourced and public), \textsf{QGym}, suitable for benchmarking queueing policies across a wide range of problem instances. Our framework can simulate systems with general, non-exponential, and non-stationary event time distributions, requiring only samples from these distributions, which may be obtained from datasets.

This is achieved through the \emph{discrete event dynamical system} representation of queuing networks, a dominant paradigm for queuing simulation (see, e.g., AnyLogic \citep{borshchev2014multi}, SimPy \citep{matloff2008introduction}). 
Although our queueing setting is markedly different, our framework is broadly inspired by OpenAI Gym~\citep{mitzenmacher2001power}. Our modeling framework helps bridge the gap between existing applications of RL, which deal with idealized environments, and industrial simulation paradigms for performance analysis in real-world systems.


\begin{figure}[t]
\vspace{-1cm}
  \centering  \includegraphics[width=1.\textwidth]
  {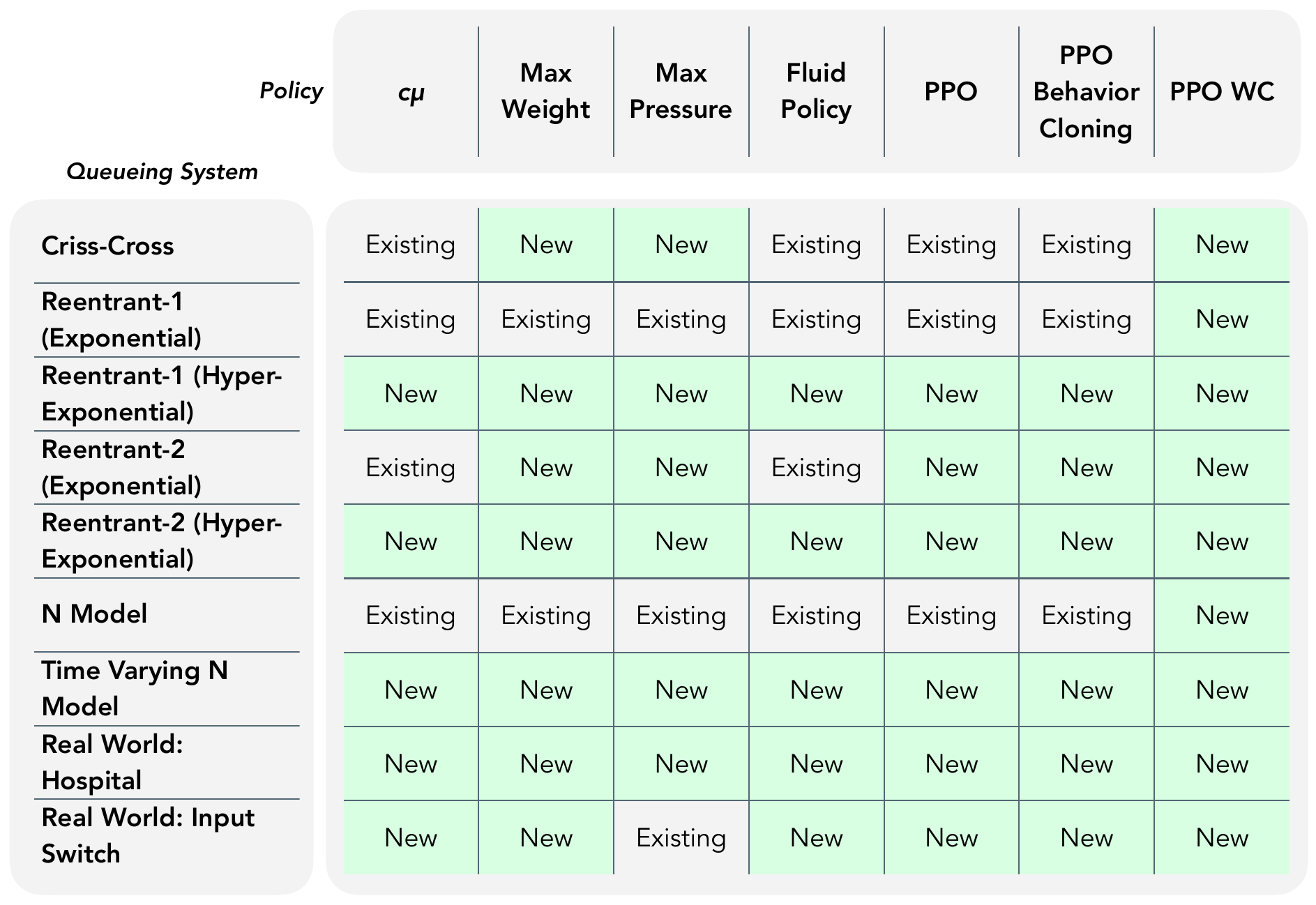}
  \caption{\textsf{QGym} provides a unified and comprehensive benchmarking system for queueing policies, across a range of realistic environments.}
  \label{fig:ingredients}
\end{figure}

We instantiate our abstract and flexible queuing simulation framework with a comprehensive list  of queueing environments.
Specifically, we consider parallel server systems motivated by skill-based routing problems in service system applications, where the processing speeds depends on both the server type and job type (match of skills)~\citep{chen2020survey}.
We also implement the criss-cross network that is widely studied in the queuing control literature \citep{martins1996heavy}.
Finally, we consider networks with tandem and reentrant structures that arise in both manufacturing and service systems and is known to suffer from bottleneck resources \citep{li2008production, huang2015control}. 

The initial set of environments we provide include systems calibrated from real-world applications. For instance, we have a parallel server system with 8 customer classes and 11 different types of servers (server pools) modeling patient flow in the hospital inpatient ward network \citep{dong2020structural}. 

Finally, we provide a comprehensive list of baseline policies that span multiple literature. From the classical queueing literature, we implement the $c\mu$-rule, the MaxWeight and maximum pressure policy, and the fluid-based policies \citep{bertsimas2014robust}.
For model-free RL algorithms, we implement several variations of PPO algorithms tailored for queuing network controls. These resources aim to facilitate a thorough and standardized evaluation of RL methods in diverse queuing scenarios.

Taken together, the \textsf{QGym} framework provides the first comprehensive and flexible framework for benchmarking queuing algorithms across a range of different environments. Our initial empirical benchmarking highlight the following considerations that impact the practical performance of RL policies (Sec. \ref{sec:Experiments}).
\begin{itemize}[leftmargin=*,itemsep=-0em]
\item \textbf{Policy architecture is important.} Without any modifications, RL algorithms such as Proximal Policy Optimization (PPO) fail to achieve stability. But equipped with a simple modification inspired by queuing theory, it is able to outperform baseline queuing methods in 77\% of instances.
\item \textbf{Performance gains of RL are larger in noisy, non-exponential environments.} Our modified PPO is able to tailor the policy to the higher noise environment, achieving larger relative gains.
\item \textbf{Larger networks are still hard to control.} PPO mostly outperforms queuing baselines in small networks, but struggles for larger, more realistic ones
\end{itemize}

\begin{wrapfigure}{r}{0.5\textwidth}
\vspace{-.3cm}
\scriptsize
\centering
\begin{tabular}{
    >{\arraybackslash}p{1.7cm} 
    >{\centering\arraybackslash}p{1.2cm} 
    >{\centering\arraybackslash}p{1.4cm} 
    >{\centering\arraybackslash}p{0.4cm} 
}
\toprule 
 & $\begin{array}{c}
     \text{Queuing + RL}\\
     \text{\cite{moallemi2008approximate, shah2020stable, qu2020scalable}} \\
     \text{\cite{dai2022queueing, liu2022rl, wei2024sample}}
 \end{array}$ 
 & $\begin{array}{c}
     \text{Simulators}\\
     \text{Simio~\cite{Simio2024}}\\
     \text{AnyLogic~\cite{AnyLogic2024}}
 \end{array}$  
 & Ours\tabularnewline
\midrule
$\begin{array}{l}\hspace{-1em}\text{Gym environment}\end{array}$ & $\textcolor{green}{\text{{\cmark}}}$ & $\textcolor{red}{\text{{\xmark}}}$ & $\textcolor{green}{\text{{\cmark}}}$\tabularnewline
\midrule 
$\begin{array}{l}\hspace{-1em}\text{Event-driven}\end{array}$ & $\textcolor{red}{\text{{\xmark}}}$ & $\textcolor{green}{\text{{\cmark}}}$ & $\textcolor{green}{\text{{\cmark}}}$\tabularnewline
\midrule 
$\begin{array}{l}
\hspace{-1em}\text{General + non-stationary}\\
\hspace{-1em}\text{event times}
\end{array}$ & $\textcolor{red}{\text{{\xmark}}}$ & $\textcolor{green}{\text{{\cmark}}}$ & $\textcolor{green}{\text{{\cmark}}}$\tabularnewline
\midrule
$\begin{array}{l}
\hspace{-1em}\text{Benchmarking + } \\
\hspace{-1em}\text{Policy Implementation}
\end{array}$ & $\textcolor{red}{\text{{\xmark}}}$ & $\textcolor{red}{\text{{\xmark}}}$ & $\textcolor{green}{\text{{\cmark}}}$\tabularnewline

\bottomrule
\end{tabular}
\vspace{-0.5em}
\caption{\footnotesize Comparison of our work with related methods}
\label{table:comparison}
\end{wrapfigure}
\paragraph{Related Work.}
Our work is related to three bodies of work. First, it connects to the research developing RL algorithms for queuing network control problems (see, \citep{moallemi2008approximate, shah2020stable, qu2020scalable, dai2022queueing, liu2022rl, wei2024sample} for some recent development). Most of these studies focus on stationary and Markovian systems, and the algorithm performance is empirically tested on a limited set of problem instances. Our work complements this research by creating a flexible queuing simulation framework that can handle more complex systems. \textsf{QGym} offers a diverse set of problems for empirically validating the performance of different RL algorithms and comparing them to standard queueing policies. 

Second, our work is related to research on discrete-event simulation \citep{fishman2001discrete, asmussen2007stochastic, nelson2010stochastic} and simulation software such as Simio and AnyLogic. Our work builds on these foundations and extends them to facilitate the benchmarking of RL methods in an \emph{open-sourced, public forum} so that the research community can build on our framework (e.g., crowdsourcing more environments).

Lastly, our work aligns with the growing efforts to build RL library and benchmarking suite for sequential decision-making problems in Operations Research \citep{hubbs2020or, archer2022orsuite, eckman2023simopt}. Our contribution complements these initiatives by focusing specifically on queuing network control problems. This specialization enables us to develop a tailored queuing simulation environment that is highly flexible and capable of addressing a diverse set of queuing control problems. See Table \ref{table:comparison} for a comparison between our work and related lines of work.

\section{An Event-Driven Queuing Simulation Framework}

We implement the following key features in order to design a flexible framework for training and evaluating queueing policies across across diverse environments. 
\begin{enumerate}
\item Event-Driven Architecture: To address the continuous time nature of the problem, \textsf{QGym} employs an event-driven approach, where system states are updated when new events occur. This enhances the scalability of our framework and allows supporting \textit{arbitrary} arrival patterns, in contrast to traditional discrete time-step-driven models. 
\item Extensive Customizability: \textsf{QGym} allows extensive customization in the queuing network topology, job processing pipelines, and stochastic inputs, enabling a broad set of queuing systems that meet the needs of both academic and industrial applications (see Sec. \ref{sec:Experiments}). 
\item OpenAI Gym Integration: Built on the OpenAI Gym interface, \textsf{QGym} facilitates easy testing and deployment of diverse queuing policies, both RL-based and traditional. Its modular design promotes easy integration of new functionalities, supporting continuous evolution.
\end{enumerate}

\paragraph{Event-Driven MDP Formulation}

We begin by describing how to convert a classical multi-class queuing scheduling problem into an \textit{event-driven} MDP problem. Consider a queuing system with $M$ queues and $N$ servers. Assume a Poisson arrival for each queue with an arrival rate $\lambda_{i}$ for $i \in [M]$. Jobs within a queue are processed on a first-come-first-served basis. A decision-maker can assign a job from queue $i$ to server $j$ if $j$ is available, at a service rate $\mu_{ij}$. Each server can serve only one job at a time and each queue can only be served by one server at a time. Each job in queue $i$ incurs a holding cost $c_{i}$ per unit of time. The objective is to design rules to match jobs with servers that minimize the total holding cost over a finite time horizon.

For the MDP formulation, consider step $k$ associated with timestamp $t_k$. Let $Q_{i}(t_k) \in \mathbb{N}$, for $i \in [M]$, represent the queue length of queue $i$ at step $k$. Let $\tau_{i}^{\rm A}(t_k) \in \mathbb{R}$, for $i \in [M]$, denote the \emph{residual inter-arrival time} at queue $i$ from time $t_k$.
Define the action $a(t_{k}) \in \{0, 1\}^{N \times M}$ to decide the assignment jobs to servers at step $k$ where $a_{ij}(t_{k}) = 1$ if a job at queue $i$ is assigned to server $j$; otherwise, it is 0, subject to feasibility. After making the decision, $\tau_i^{\rm S}(t_k) /\mu_{ij}\in \mathbb{R}$ is the \emph{residual service time} for queue $i$ if server $j$ is assigned (the time is infinite if the job is not assigned to a server). To express this compactly, we define $\mu_{i}(t_{k}) \equiv \mu_{i}^{\top} a_{i}(t_{k})$, with $\mu_{i}$ and $a_{i}(t_{k})$ being the $i$th rows, and refer to the service time as $\tau_i^{\rm S}(t_k) / \mu_{i}(t_{k})$, which will be equal to $\infty$ if no server is assigned to queue $i$. The state of the MDP is the tuple $s_{k} = (t_{k}, Q(t_{k}), \tau^{A}(t_{k}), \tau^{S}(t_{k}))$.

It is worth emphasizing that almost always the controller only observes the queue-length $Q(t_{k})$. Considering that our framework involves an expanded state space that includes residual event times, our framework is technically a \emph{partially observable Markov Decision Process} (POMDP), with observations $o_{k}$ only consisting of the current time-stamp $t_{k}$ and the queue-lengths $Q(t_{k})$,
\[
o_{k} = (t_{k}, Q(t_{k})), \quad \quad s_{k} = (t_{k}, Q(t_{k}), \tau^{A}(t_{k}), \tau^{S}(t_{k})).
\]

The next event is the event with the minimum remaining processing time, which we denote as the inter-epoch time $\tau_{k+1}$,
\[
\tau_{k+1} = \min \left\{\tau_i^{\rm A}(t_k), \frac{\tau_{i}^{S}(t_{k})}{\mu_{i}(t_{k})} | i \in [M]\right\}
\]

The event determines the update to the state. If an arrival to queue $i$ occurs, then the queue-length $Q(t_{k+1})$ is incremented by 1 in the $i$th position which is represented by adding the canonical basis vector $e_{i}$. If a service occurs, then the queue-length $Q(t_{k+1})$ will be updated by the vector $\Delta_{i} \in \{1,0,-1\}^{N}$. Typically, this update will involve decrementing the queue-length in the $i$th position by 1 since a job departs from queue $i$. At the same time, this could increment the queue-length of another queue, as is the case for tandem queues where jobs are routed to other queues after processing.
\begin{align}
Q(t_{k+1}) &= Q(t_{k}) 
+ \sum_{i=1}^{N} e_{i}
\mathbf{1}\{ \tau_{i}^{A}(t_{k}) = \tau_{k+1} \} 
+ \sum_{i=1}^{N} \Delta_{i}
\mathbf{1}\left\{ \frac{\tau_{i}^{S}(t_{k})}{\mu_{i}(t_{k})}  = \tau_{k+1} \right\}.
\end{align}

After the event occurs, the residual event times are reduced by $\tau_{k+1}$, which is the time that elapsed between events. For the event which occurred, the residual event time are reset by drawing new event times from a distribution. We denote the new event times as  $T_{i}^{\rm A}$ if the event was an arrival and $T_{i}^{\rm S}$ if the event was a service,
\begin{align}
\tau_{i}^{\rm A}(t_{k+1}) &= 
\tau_{i}^{\rm A}(t_{k}) - \tau_{k+1} +
T_{i}^{A} \cdot \mathbf{1}\{ \tau_{i}^{A} = \tau_{k+1} \} \\
\tau_{i}^{\rm S}(t_{k+1}) &= 
\tau_{i}^{\rm S}(t_{k}) - \mu_{i}(t_{k})\tau_{k+1} +
T_{i}^{S} \cdot \mathbf{1}\left\{ \frac{\tau_{i}^{S}(t_{k})}{\mu_{i}(t_{k})}  = \tau_{k+1} \right\}
+ \infty \cdot \mathbf{1}\{ Q(t_{k+1}) = 0\}.
\end{align}
Since services cannot occur to an empty queue, the service time is infinite if the queue-length is zero.


This event-driven MDP formulation advances time steps by events, i.e., the next event time, in contrast to existing RL work in queuing that often uses a time-step-driven MDP which poses scalability challenges in real-world applications \citep{menda2018deep}.

\paragraph{Functionality}
We implement functionalities to generalize the classical multi-class, multi-server problem described above. This enables \textsf{QGym} to accommodate an extensive range of queuing problems in real-world applications (e.g., see the hospital example in Sec.\ 4). While not exhaustive, our system is designed to be flexible, allowing new models to be easily incorporated and tested.

 \emph{Network Topology.} We allow customized network topologies. Given a binary matrix \(B \in \mathbb{R}^{M \times N}\), server \(j\) is permitted to serve jobs from queue \(i\) only when \(B_{ij} = 1\).

\emph{Job Transition.} Completed jobs from queue \(i \in [M]\) can transition to another queue \(i' \in [M]\) instead of leaving the system. This facilitates complex job processing pipelines, such as re-entrant and tandem queues.

\emph{Arbitrary Arrivals.} Users can define arbitrary arrival patterns for queues via a Python function that inputs the current time and outputs the time until the next arrival. This feature allows for the simulation of time-varying and non-Poisson arrivals. Arrivals ``generated" from real data are also supported.

\emph{Service Time Distribution.} Service times are drawn from arbitrary distributions specified by the user, with service rates as parameters. Although time-varying service rates are not yet supported, they can be implemented in a manner similar to arrival patterns.

\emph{Server Pool.} Users can define each class of server as a pool. When many servers share the same characteristics, users can specify the number of servers in each class (server pool) instead of creating numerous separate servers and inflating the size of the network matrix. This mechanism enables the simulation of large-scale systems without compromising performance.

\emph{Job-Level Tracking.} The simulator tracks states at the job level, monitoring the service time for each job in a queue. The fine-grained job-level tracking enables simulation of parallel-server systems, where multiple servers can serve a single queue. To illustrate why a less fine-grained choice of tracking at queue level could fail, we consider the cases where multiple servers serve a single queue and the jobs are preempted. Only tracking at queue level does not allow recording remaining service time for each job and resuming service them later. Tracking states at job level enables flexible simulations of parallel server systems.


\emph{Reward.} Users can define rewards using arbitrary functions on states and actions. In this paper, we focus on minimizing holding costs as a representative example.

\begin{figure*}[t]
  \centering
  \includegraphics[width=0.7\textwidth]{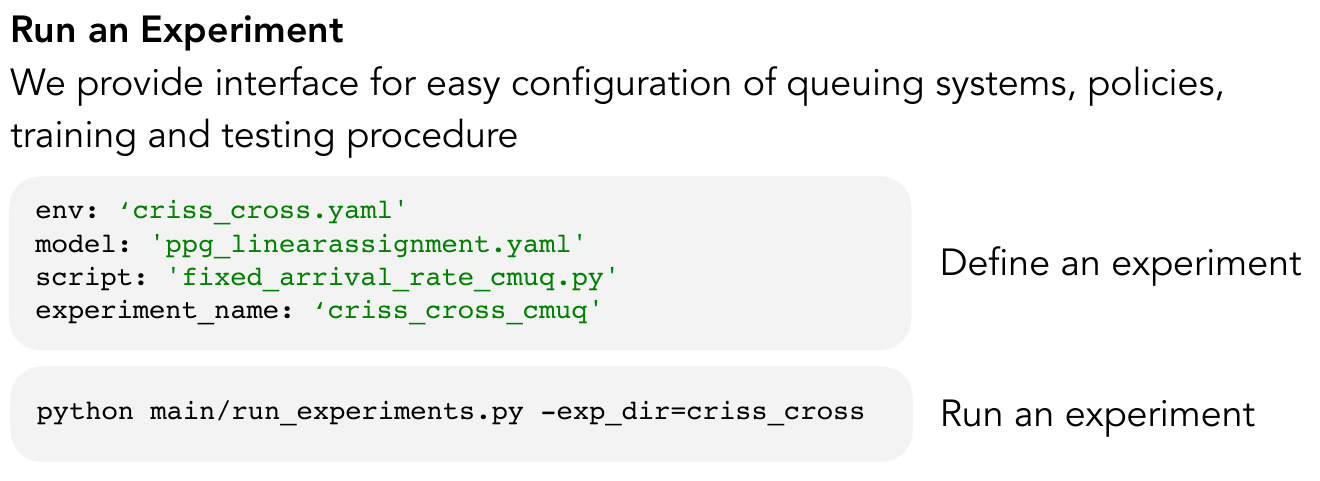}
  \caption{\textsf{QGym} provides an user-friendly interface to define and run experiments for evaluating routing policies on queuing networks.}
  \label{fig:run_experiment}
\end{figure*}

\vspace{-10pt}
\paragraph{OpenAI Gym Design.}
The simulator environment is structured as an OpenAI Gym environment, adhering to its design principles. This allows users to train and test a variety of reinforcement learning algorithms seamlessly. Each simulation trajectory consists of a sequence of steps (defined in the OpenAI Gym \texttt{step} format) and supports batch-based GPU execution to accelerate computation. We provide a range of environments (e.g., N-model, reentrant, re-reentrant, criss-cross, etc.) and policies, including both RL methods and traditional ones (see Sec.\ 3), to facilitate easy testing. Users can conveniently configure environment and policy parameters using \texttt{.yaml} files, and new environments and policies can be added by following the OpenAI Gym convention. See Figure \ref{fig:run_experiment} for code snippets of our user-friendly interface for defining and runnning experiments. More details can be found in the Appendix \ref{sec:simulator-design}. 

\section{Benchmark Policies}
In this section, we introduce the queuing policies benchmarked in our testbed. Formally, each policy $\pi(\cdot|o)$ maps an observation $o \in \mathbb{N}^{N}$ (the current queue-lengths)  to an action $a \in \{0,1\}^{N\times M}$. These include both traditional control-based policies and RL-based policies tailored for queuing.

\subsection{Traditional Policies}
 All traditional policies considered fall into the class of policies using the linear assignment rule: given a policy $\pi$, a priority matrix $\rho \in \mathbb{R}^{M \times N}$ is outputted from $\pi$ at each step, the action (i.e., job-server assignment) is then decided by 
 $$\max_{a\in\mathcal{A}}\sum_{i,j} \rho_{ij} a_{ij}$$
where $\mathcal{A} \subset \mathbb{R}^{M\times N}$ captures the feasibility (e.g., compatibility and resource capacity) constraints. 

\textbf{\(c\mu\)-rule.} A classic policy for scheduling multiple classes of jobs is the \(c\mu\)-rule, which has been shown to minimize the linear waiting cost in multi-class single-server queues \citep{cox1961queues}. In this case, $\rho_{ij}=c_i \mu_{ij}$. Server $j$ prioritizes the queue with a larger $c_i\mu_{ij}$-index, where \(c_i\) denotes the holding cost per job per unit time for queue \(i\), and \(\mu_{ij}\) is the service rate when server $j$ processes a job from queue $i$. 


\textbf{MaxWeight.} Another important class of policies is known as MaxWeight policies, which has been shown to be maximally stable for single-hop networks \citep{tassiulas1990stability} and are also known for their favorable asymptotic properties under a resource pooling condition  \citep{stolyar2004maxweight}. We consider a specific form of MaxWeight policy where $\rho_{ij} = c_{i}Q_{i}\mu_{ij}$ with $Q_i$ being the queue length of queue \(i\). Here, by taking the queue lengths into account, we are able to better balance the workload in the system.

\textbf{Maximum pressure.} The maximum pressure policies, which are also known as the back pressure policies, are similar to the MaxWeight policies but account for workload externality within the network. This additional consideration allows for better workload balancing in the multihop setting, especially in networks with tandem or reentrant structures. These policies have been shown to be maximally stable in multi-hop networks \citep{tassiulas1990stability, dai2005maximum}. We consider a specific form of the maximum pressure policy under which $\rho_{ij}= (c_{i}Q_{i}\mu_{ij}-\sum_{k=1}^{M}c_kQ_k\mu_{ij}p_{ik})$, where $p_{ik}$ is the probability that after a class $i$ job is processed by server $j$, it will join queue $k$ next. Note that when $p_{ik}=0$ for all $k$, the maximum pressure policy simplifies to the MaxWeight policy. However, when $p_{ik}>0$ for some $k$, the maximum pressure policy accounts for the fact that processing a class $i$ job will generate a class $k$ job, thus considering the impact on ``downstream" queues.

\textbf{Fluid Policy.} One can derive a `fluid model' of the queuing network as a system of ordinary differential equations (ODEs) driven by the service and arrival rates of the network~\cite{chen1993dynamic}. By discretizing the ODEs on a finite grid, one can minimize the linear holding costs by solving a linear program (LP). We then use the computed priorities $\rho_{ij}$ in the original queuing network. To maintain fidelity with the original dynamics, we periodically re-solve the LP. We solve the LP via CVX~\cite{diamond2016cvxpy}, and resolve after every 1000 steps.



\subsection{Deep RL based methods}
Proximal Policy Optimization (PPO) has been a popular choice for applying RL to queuing  \cite{liu2022rl,dai2022queueing}. Following the convention, we implemented a few variants of PPO in our testbed. We apply existing and develop new modifications to improve the stability and scalability of PPO in queuing systems. 

\textbf{PPO.} The action space in our problem is \(a \in \mathbb{R}^{M\times N}\). Thus, directly applying vanilla PPO \cite{schulman2017proximal} will suffer from the explosion of dimensionality. To address this issue, we require \(\{a_{ij}\}_{i=1}^{M}\) to be a probability distribution for all \(j \in [N]\). We sample \(a_{j}' \sim \{a_{ij}\}_{i=1}^{M}\) for each \(j\) independently to decide which queue server \(j\) serves. The feasibility constraint is then verified by the environment.
Compared to the existing queuing RL method that discretizes the action space \cite{dai2022queueing}, this parameterization is much more scalable when \(M\) and \(N\) both grow. The classic normalization tricks have been implemented to make PPO more stable (e.g., advantage function normalization, reward normalization, etc.). In addition, to reduce the variance for the Generalized Advantage Estimation (GAE) in PPO
$$\hat{A}_t^{\mathrm{GAE}(\gamma, \lambda)} := \sum_{l=0}^{T}(\gamma \lambda)^l \delta_{t+l}^V ~~\mbox{where}~~ \delta_t^V=r_t+\gamma V\left(s_{t+1}\right)-V\left(s_t\right),$$
we truncate \(T\) to \(T_0\) where \(T_0\) is the first time that all queues are empty (i.e., regenerative point).

\textbf{PPO with Behavior Cloning (PPO-BC).} Implementing the PPO described above with a random initial policy still suffers from poor performance due to instability. Similar to \cite{dai2022queueing}, we address this by using behavior cloning. We first train \(\pi\) to imitate a Max-Weight style policy that assigns servers to classes with probability proportional to $e^{Q_{i}}$.
Using this procedure as a warm-up significantly enhances the stability of training and achieves much better results compared to PPO alone. 

\textbf{Work-Conserving PPO (PPO-WC).} We have a simple observation: the policy should never assign server capacity to an empty queue (so-called `work conservation' rule \cite{dai1995stability}). We impose this `inductive bias' to the policy design directly. To do so, we mask the probabilities $a_{ij}$ obtained from PPO while preserving differentiability:
$$
a_{ij} = \frac{a_{ij} \mathbf{1} \{Q_i >0\} }{\sum_{i=1}^{M} a_{ij} \mathbf{1} \{Q_i >0\}}
$$

where $Q_{i}$ is the length of queue $i$ when taking actions. In case the queues are all empty, we avoid division-by-zero errors by clipping the denominator for some small $\epsilon$. As we observe in the experiments in Section~\ref{sec:Experiments}, this small change greatly improves the performance of PPO. With randomly initialized policy parameters, training algorithms utilizing WC policy parameterization consistently outperform those using vanilla parameterization. Notably, the PPO-WC training algorithm demonstrates a high training stability, such that action clipping is almost never required, which was the core advantage of PPO. To further validate the effectiveness and advantages of WC parameterization, we implemented \textbf{A2C} (a vanilla actor-critic algorithm without clipping and KL regularization) with the same WC parameterization and observed comparable performance to that of PPO-WC. These results underscore the robustness and generalizability of WC parameterization.

\section{Experiments}\label{sec:Experiments}

Using our environment, we benchmark the performance of PPO and traditional queuing baselines across a diverse suite of queuing networks. We curate a set of queuing network instances, drawing upon networks studied in the queuing literature as well as novel instances, with coverage of network architectures relevant to manufacturing, hospital patient flow, and wireless network applications. Overall, we observe that while PPO alone performs quite poorly, \textit{PPO-WC outperforms the traditional policies in 77\% of all instances}, highlighting the importance of incorporating queuing structure in  RL policy design.



\subsection{Setup}
\begin{figure}
    \vspace{-1.8cm}
  \centering
  \includegraphics[width=1.\textwidth]{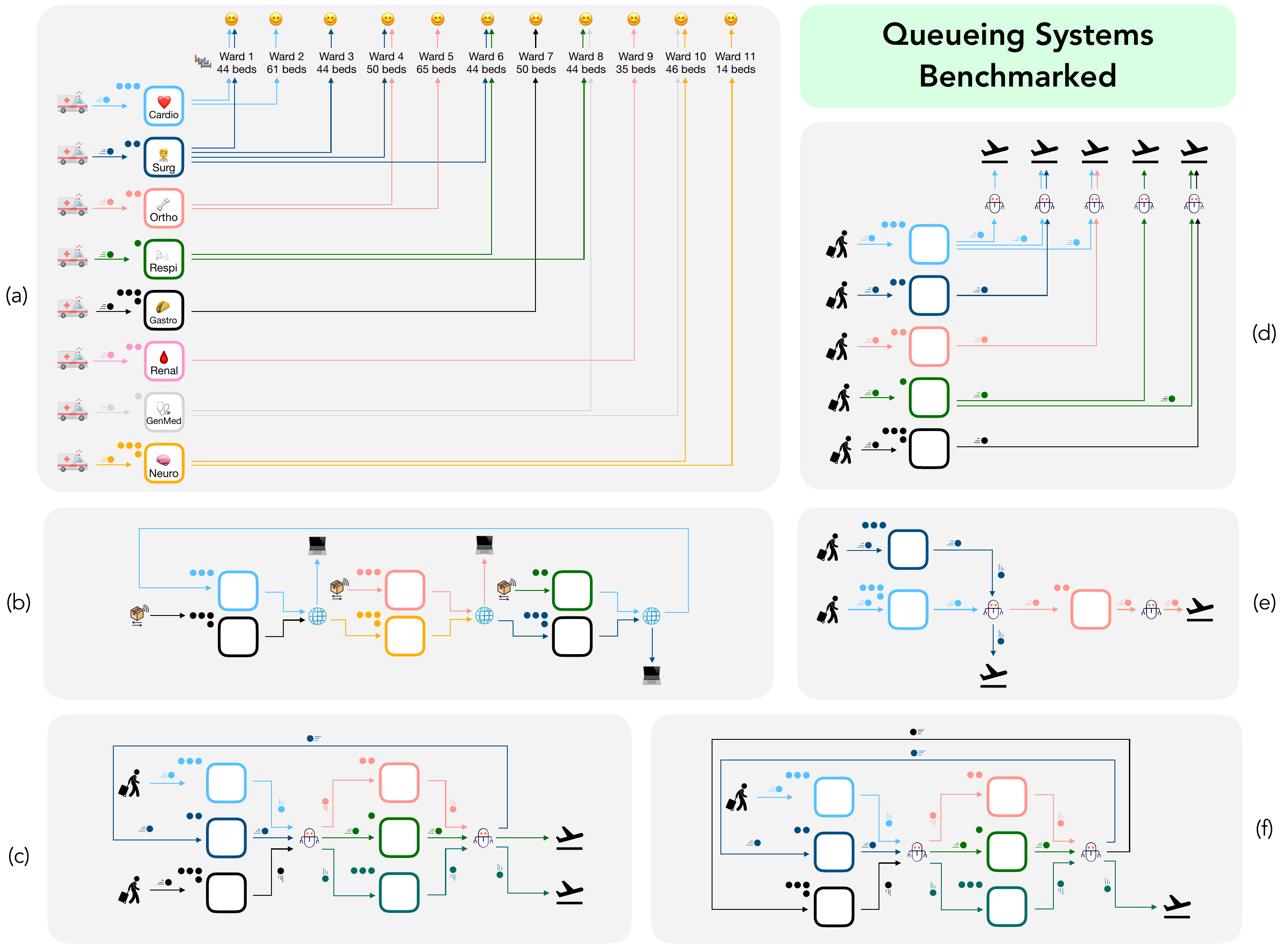}
  \caption{Queuing systems \textsf{QGym} benchmarks. (a) Real-world example of hospital routing. (b) Real-world example of data input-switch routing. (c) and (f) Two variations of reentrant networks. (d) Five-by-Five netowrk for modeling call centers. (e) Criss-cross network. See details in section \ref{sec:Experiments}}
  \label{fig:network}
\vspace{-1em}
\end{figure}

\textbf{Network structure.} In total, we consider 20 unique problem instances across the following networks. See Fig.~\ref{fig:network} for the corresponding network topologies. 
\begin{enumerate}[label=(\alph*),leftmargin=*]
\item {\bf Hospital}: Patients arrive to $M = 8$ specialties (Cardiology, Surgery, Orthopedics,  Respiratory disease, Gastroenterology and endoscopy, Renal disease, General Medicine, Neurology) split across $11$ inpatient wards. Each ward consists of multiple beds (servers). In total, this is modeled by $N = 497$ servers across the $11$ wards. The hospital employs a focused-care model where each ward is primarily designated to serve patients from one specialty or two specialties. 
The network topology, arrival rates, and service rates are calibrated to a real hospital setting. 
\item {\bf Input-Queue Switch}~\cite{dai2005maximum, mckeown1999islip}: Packets in a crossbar switch arrive to $M = 6$ queues, and are processed by $N = 3$ servers. 
\item {\bf Reentrant ($L$)}~\cite{bertsimas2014robust,dai2022queueing}: Manufacturing lines process goods in several sequential steps $L$. We consider a family of instances with $L \in \{2,...,10\}$ with $M = 3L$ queues and $N = L$ servers. High variance in the service times can lead to bottlenecks in the network, and so we also consider instances with hyper-exponential service times, which are mixtures of exponential distributions.
\item {\bf Five-by-Five Network}~\cite{chen2023optimal}: Call-centers route customers from $M = 5$ classes to $N = 5$ servers. Call center demand changes throughout the day, which we model through time-varying inter-arrival times $\tau^{A} \sim \text{Exp}(\lambda(t))$.
\item {\bf Criss-Cross}~\cite{harrison1990scheduling, avram1995fluid, martins1996heavy}: A standard reentrant network considered in the literature, consisting of $M = 3$ queues and $N = 2$ servers. 
\item {\bf N-model}~\cite{harrison1998heavy}: A standard parallel-server system considered in the literature, consisting of $M = 2$ queues and $N = 2$ servers. 

\end{enumerate}

\textbf{Objective.} The core performance metric we consider is the long-run-average total queue-length, which is approximated by averaging over a long horizon of $n$ events.
\begin{equation}
  \mathbb{E} \left[\frac{1}{t_n}\sum_{k=1}^{n} \sum_{i=1}^{N} Q_{i}(t_{k})(t_{k+1} - t_{k}) \right]
=  \mathbb{E} \left[\frac{1}{t_{n}} \int^{t_{n}}_{0} \sum_{i=1}^{N} Q_{i}(t)dt \right]
\end{equation}
For each policy, we estimate the expected time-average total queue length by evaluating the policy over $100$ trajectories. We also report the corresponding standard errors.

\textbf{Training Procedure}
All PPO variants were trained under the same conditions. Each policy was trained over $100$ episodes, each consisting of $50,000$ environment steps parallelized over $50$ actors. Following existing works, we used a discount factor of $\gamma = 0.998$, a GAE parameter of $\lambda=0.99$, and set the KL divergence penalty of $\beta=0.03$. For the value network, we used a batch size of $2500$, while for the policy network, we used the entire rollout buffer (batch size of $50,000$) to take one gradient step. We performed $3$ PPO gradient updates on the same rollout data. For all the experiments, we used Adam optimizer with a cosine decaying warming-up learning rate scheduler. The learning rates were set to $3 \times 10^{-4}$ for the value network and $9 \times 10^{-4}$ for the policy network. We used 3\% of the training horizon to warm up to the maximum learning rate and then cosine decayed to $1 \times 10^{-5}$ for both networks.

\subsection{Results}
Tables 1-5 document the time-averaged total queue length for the policies we consider. Our systematic benchmarking
illustrates the differences in practical performance of reinforcement learning and traditional queuing policies. Our findings can be summarized into three folds. 

\textbf{PPO learns an effective controller, but only under the right policy architecture.} Without any modifications, in every setting PPO fails to stabilize the queuing network, systematically confirming an observation made in previous works~\cite{liu2022rl}. Behavior cloning a stabilizing policy drastically improves the training process, yet the policy still fails to achieve parity with the traditional queuing methods. It is only when we endow PPO with a work-conserving policy, that we are able to completely stabilize the training process and surpass the performance of traditional queuing methods in most settings (the pair-wise win rate is 77\%). This suggests that the usual `tabula rasa'~\cite{agarwal2022reincarnating} approach of reinforcement learning, which aims to learn from a completely flexible and unstructured policy class, is unsuitable for queuing control. The `inductive bias' of work-conservation not only speeds up learning, but is decisive in enabling the algorithm to improve upon existing policies.

\textbf{Performance gains from PPO are larger in noisier, non-exponential environments.} We compare the results in Table 2, which details performance for the reentrant networks under exponentially-distributed event times with Table 3, which involves hyper-exponential noise. In Table 2, the improvements of PPO-WC over other baselines are relatively modest, around a 10\% improvement at most. Yet, Table 3 shows that under hyper-exponential noise, the improvements can be larger, resulting in a relative reduction in holding costs of around 21\%  with $L = 2$. We are only able to observe these improvements because our simulation framework can incorporate general noise inputs.

\textbf{There is plenty of room for improvement.} PPO-WC mainly outperforms queuing baselines on small-scale examples, but incurs a higher cost in more realistic, larger-scale instances such as the hospital network, despite being trained over 5 million steps. This gap in performance points towards the sample-inefficiency of RL in larger networks.

Additional experiment results and a detailed setup for reproducibility is provided in Appendix \ref{sec:additional-benchmark}, \ref{sec:code-release}, and \ref{sec:comp-resources}.

\section{Conclusion}
We propose a simulation framework and a comprehensive suite of benchmarks for queuing network control. Although RL is capable of training effective controllers, there still remain performance gaps compared to standard baselines and we hope that our simulation framework can enable algorithmic progress towards bridging these gaps.


\vspace*{\fill}
\begin{table}[t]

\scriptsize
\caption{Criss Cross}
\label{table:criss-cross}
\hspace{-1.5cm}
\begin{tabular}{|l|c|c|c|c|c|c|c|c|}
\hline
Network & $c\mu$ & MW & MP & FP & PPO & PPO BC & PPO WC & A2C WC\\ \hline
Criss Cross BH & $16.1 \pm 0.3$  & $\mathbf{15.3 \pm 0.3}$  & $ 19.0 \pm 0.3$ & $18.2 \pm 2.7$ & $8.6\text{E+}3 \pm 4.6$ & $24.0 \pm 0.2$ & $\mathbf{15.4 \pm 0.2}$ & $15.29 \pm 0.2$ \\ \hline
\end{tabular}

\label{table:reentrant-1}
\scriptsize
\vspace{1em}
\caption{Reentrant [Exponential]}
\hspace{-1.5cm}
\begin{tabular}{|l|c|c|c|c|c|c|c|c|}
\hline
$L$ & $c\mu$ & MW & MP & FP & PPO & PPO BC & PPO WC & A2C WC\\ \hline
2 & $19.0 \pm 0.4$ & $14.8 \pm 0.4$ & $18.9 \pm 0.5$ & $16.8 \pm 4.3$ & $1.8\text{E+}3 \pm 6.6$ & $25.1 \pm 0.6$ & $13.6 \pm 0.4$ & $\mathbf{13.01 \pm 0.3}$\\ \hline
3 & $\mathbf{21.6 \pm 0.6}$ & $24.8 \pm 0.7$ & $30.9 \pm 1.0$ & $27.7 \pm 4.4$ & $1.0\text{E+}4 \pm 21.6$ & $48.2 \pm 0.5$ & $22.6 \pm 0.4$ & $22.0 \pm 0.3$\\ \hline
4 & $30.1 \pm 0.9$ & $32.1 \pm 1.1$ & $40.3 \pm 1.4$ & $40.6 \pm 4.8$ & $1.8\text{E+}4 \pm 41.7$ & $183.4 \pm 5.2$ & $\mathbf{29.7 \pm 0.4}$ & $30.7 \pm 0.4$\\ \hline
5 & $51.3 \pm 1.3$ & $50.0 \pm 1.5$ & $52.2 \pm 1.6$ & $49.8 \pm 5.1$ & $2.7\text{E+}4 \pm 84.4$ & $135.2 \pm 3.1$ & $\mathbf{38.7 \pm 0.4}$ & $39.1 \pm 0.5$\\ \hline
6 & $54.7 \pm 1.6$ & $49.2 \pm 1.3$ & $59.1 \pm 2.1$ & $ 54.5 \pm 4.2$ & $5.9\text{E+}4 \pm 315.3$ & $358.0 \pm 9.7$ & $48.5 \pm 0.5$ & $\mathbf{47.4 \pm 0.5}$\\ \hline
7 & $56.4 \pm 1.6$ & $\mathbf{54.4 \pm 1.8}$ & $70.5 \pm 2.9$ & $63.7 \pm 6.7$ & $4.4\text{E+}4 \pm 208.1$ & $526.6 \pm 8.7$ & $56.3 \pm 8.2$ & $56.7 \pm 0.8$\\ \hline
8 & $\mathbf{59.4 \pm 2.2}$ & $68.0 \pm 1.7$ & $81.4 \pm 3.0$ & $ 74.0 \pm 6.7$ & $5.9\text{E+}4 \pm 315.2$ & $868.5 \pm 6.0$ & $65.8 \pm 6.2$ & $67.5 \pm 0.6$\\ \hline
9 & $72.7 \pm 2.5$ & $\mathbf{64.4 \pm 2.1}$ & $90.8 \pm 3.2$ & $83.1 \pm 7.1$ & $1.1\text{E+}5 \pm 2219.7$ & $1304.5 \pm 10.1$ & $75.8 \pm 0.7$ & $77.0 \pm 0.8$\\ \hline
10 & $87.7 \pm 2.7$ & $\mathbf{80.1 \pm 1.9}$ & $100.5 \pm 3.2$ & $93.6 \pm 8.0$ & $1.6\text{E+}5 \pm 852.8$ & $3809.1 \pm 10.4$ & $83.1 \pm 0.7$ & $90.2 \pm 0.9$\\ \hline
\end{tabular}
\label{table:reentrant_1_hyper}
\scriptsize
\vspace{1em}
\caption{Reentrant [Hyperexponential]}
\hspace{-1.5cm}
\begin{tabular}{|c|c|c|c|c|c|c|c|c|}
\hline
L & $c\mu$ & MW & MP & FP & PPO & PPO BC & PPO WC & A2C WC \\
\hline
2 & $31.69 \pm 1.3$ & $\mathbf{22.40 \pm 1.2}$ & $43.8 \pm 1.8$ & $43.6 \pm 7.5$ & $9.9\text{E+}3 \pm 20.7$ & $62.7 \pm 1.3$ & $29.9 \pm 0.7$ & $30.2 \pm 0.7$\\
\hline
3 & $\mathbf{36.76 \pm 1.9}$ & $43.00 \pm 2.2$ & $68.7 \pm 2.7$ & $59.2 \pm 8.2$ & $19.6\text{E+}3  \pm 58.0$ & $305.1 \pm 13.8$ & $47.5 \pm 0.8$ & $47.8 \pm 1.1$\\
\hline
4 & $\mathbf{58.58 \pm 2.5}$ & $74.54 \pm 2.8$ & $89.4 \pm 3.6$ & $75.6 \pm 15.3$ & $18.9\text{E+}3 \pm 53.1$ & $167.2 \pm 5.1$ & $64.4 \pm 1.2$ & $62.8 \pm 1.4$\\
\hline
5 & $\mathbf{68.91 \pm 4.0}$ & $73.19 \pm 3.7$ & $112.0 \pm 4.9$ & $97.0 \pm 12.9$ & $48.0\text{E+}3 \pm 153.5$ & $913.4 \pm 19.9$ & $81.8 \pm 1.1$ & $84.9 \pm 1.2$\\
\hline
6 & $\mathbf{85.16 \pm 4.7}$ & $98.75 \pm 3.9$ & $126.7 \pm 6.2$ & $111.2 \pm 14.4$ & $59.1\text{E+}3 \pm 336.4$ & $2383.0 \pm 15.2$ & $99.8 \pm 1.5$ & $100.8 \pm 1.4$\\
\hline
7 & $\mathbf{100.24 \pm 5.9}$ & $119.01 \pm 3.9$ & $152.3 \pm 6.6$ & $151.0 \pm 21.3$ & $65.4\text{E+}3 \pm 325.9$ & $3054.6 \pm 16.6$ & $118.2 \pm 2.0$ & $120.5 \pm 2.1$\\
\hline

\end{tabular}

\scriptsize
\vspace{1em}
\caption{Parallel Server}
\hspace{-1.5cm}
\begin{tabular}{|l|c|c|c|c|c|c|c|c|}
\hline
Network & $c\mu$ & MW & MP & FP & PPO & PPO BC & PPO WC &A2C WC\\ 
\hline
N Model & $1.7\text{E+}2 \pm 12.3$  & $\mathbf{40.2 \pm 2.2}$  &  $\mathbf{40.2 \pm 2.2}$ & $7.9\text{E+}2 \pm 18.8$ & $8.8\text{E+}3 \pm 28.5$ & $100.9 \pm 1.9$ & $44.3 \pm 1.8$ &$49.8 \pm 0.2$\\ 
\hline
Five-by-Five & $17.8 \pm 1.2$   & $\mathbf{15.2 \pm 0.7}$  &  $\mathbf{15.2 \pm 0.7}$ & $26.7 \pm 2.5$ & $1.2\text{E+}4 \pm 15.3$ & $25.2 \pm 0.1$ & $16.8 \pm 0.2$ & $162.88 \pm 4.8$\\ 
\hline
\end{tabular}


\scriptsize
\vspace{1em}
\caption{Real World Example}
\hspace{-1.9cm}
\begin{tabular}{|l|c|c|c|c|c|c|c|c|}
\hline
Network & $c\mu$ & MW & MP & FP & PPO & PPO BC & PPO WC & A2C WC\\ \hline
Input Switch & $\mathbf{5.3 \pm 0.0}$  & $5.6 \pm 0.0$  & $4.9\text{E+}3 \pm 18.9$  & $4.9\text{E+}3 \pm 12.2$ & $7.3\text{E+}3 \pm 8.9$ & $11.8 \pm 0.1$ & $\mathbf{5.3 \pm 0.3}$ & $5.4 \pm 0.0$\\ \hline
Hospital & $\mathbf{4.4\text{E+}2 \pm 0.0}$  & $\mathbf{4.4\text{E+}2 \pm 8.9}$  & $\mathbf{4.4\text{E+}2 \pm 8.9}$  & $1.5\text{E+}3 \pm 4.9$ & $2.4\text{E+}4 \pm 6.6$ & $2.5\text{E+}4 \pm 7.0$ & $2.3\text{E+}4 \pm 7.4$ & $2.3\text{E+}4 \pm 7.3$\\ \hline
\end{tabular}
\end{table}




\bibliography{./bib}
\bibliographystyle{apalike}

\newpage

\appendix

\section{Code release}
\label{sec:code-release}
Our code is available at: \url{https://github.com/namkoong-lab/QGym}

\section{Additional Benchmark Result}
\label{sec:additional-benchmark}

We provide the benchmark results for reentrant-2 [exponential] and reentrant-2 [hyperexpenential] queuing systems in Table \ref{tab:reentrant-2-exp} and \ref{tab:reentrant-2-hyper}.

\begin{table}[htbp]

\scriptsize
\caption{Reentrant-2[Exp]}
\hspace{-1.5cm}
\begin{tabular}{|l|c|c|c|c|c|c|c|}
\hline
Network & $c\mu$ & MW & MP & FP & PPO & PPO BC & PPO WC\\ \hline
2 & $26.01 \pm 0.00$ & $17.45 \pm 0.00$ &$24.5 \pm 0.00$ & $16.7\pm0.00$ & $9.04\text{E+}3 \pm 41.13$ & $39.16 \pm 0.37$ & $\mathbf{13.72 \pm 0.22}$ \\ \hline
3 & $26.27 \pm 0.00$ & $26.65 \pm 0.00$ & $30.0 \pm 0.00$ & $61.7\pm 0.00$ & $1.82\text{E+}4 \pm 37.23$ & $48.59 \pm 0.56$ & $\mathbf{22.09 \pm 0.29}$ \\ \hline
4 & $27.62 \pm 0.00$ & $34.10 \pm 0.00$ & $41.3 \pm 0.00$ & $74.6 \pm 0.00$ & $1.77\text{E+}4 \pm 52.30$ & $79.20 \pm 1.06$ & $\mathbf{29.90 \pm 0.47}$ \\ \hline
5 & $44.82 \pm 0.00$ & $40.34 \pm 0.00$ & $49.0 \pm 0.00$ & $85.6 \pm 0.00$ & $2.58\text{E+}4 \pm 63.54$ & $91.65 \pm 1.37$ & $\mathbf{38.01 \pm 0.52}$ \\ \hline
6 & $\mathbf{54.48 \pm 0.00}$ & $46.63 \pm 0.00$ & $58.8 \pm 0.00$ & $92.5 \pm 0.00$ & $4.02\text{E+}4 \pm 349.64$ & $526.57 \pm 8.74$ & $46.80 \pm 0.47$ \\ \hline
7 & $70.25 \pm 0.00$ & $77.93 \pm 0.00$ & $68.0 \pm 0.00$ & $102.1 \pm 0.00$ & $5.78\text{E+}4 \pm 116.92$ & $352.02 \pm 6.68$ & $\mathbf{55.51 \pm 0.57}$ \\ \hline
8 & $70.32 \pm 0.00$ & $72.96 \pm 0.00$ & $77.7 \pm 0.00$ & $103.3 \pm 0.00$ & $4.79\text{E+}4 \pm 208.70$ & $1332.68 \pm 7.82$ & $\mathbf{63.15 \pm 0.70}$ \\ \hline
9 & $\mathbf{65.80 \pm 0.00}$ & $77.34 \pm 0.00$ & $84.3 \pm 0.00$ & $106.7 \pm 0.00$ & $6.54\text{E+}4 \pm 491.49$ & $1574.86 \pm 9.34$ & $70.30 \pm 0.86$ \\ \hline
10 & $81.35 \pm 0.00$ & $\mathbf{82.00 \pm 0.00}$ & $92.7 \pm 0.00$ & $120.9 \pm 0.00$ & $8.11\text{E+}4 \pm 355.34$ & $1876.54 \pm 89.20$ & $80.36 \pm 0.79$ \\ \hline
\end{tabular}
\label{tab:reentrant-2-exp}
\end{table}

\begin{table}[htbp]

\scriptsize
\caption{Reentrant-2[Hyper]}
\label{tab:reentrant-2-hyper}
\hspace{-1.5cm}
\begin{tabular}{|l|c|c|c|c|c|c|c|}
\hline
Net & $c\mu$ & MW & MP & FP & PPO & PPO BC & PPO WC\\ \hline
2 & $28.43 \pm 1.37$ & $\mathbf{22.82 \pm 1.89}$ & $58.63 \pm 2.26$ & $39.75 \pm 7.29$ & $9.75\text{E+}3 \pm 59.58$ & $66.87 \pm 1.11$ & $30.67 \pm 0.83$ \\ \hline
3 & $41.46 \pm 2.14$ & $\mathbf{36.97 \pm 2.69}$ & $67.14 \pm 2.94$ & $55.52 \pm 9.38$ & $2.05\text{E+}4 \pm 132.63$ & $482.64 \pm 17.98$ & $45.66 \pm 0.84$ \\ \hline
4 & $72.87 \pm 2.68$ & $92.73 \pm 3.68$ & $92.96 \pm 4.01$ & $72.09 \pm 14.37$ & $1.93\text{E+}4 \pm 62.07$ & $149.23 \pm 3.17$ & $\mathbf{61.09 \pm 1.30}$ \\ \hline
5 & $\mathbf{58.65 \pm 2.66}$ & $72.48\pm 3.92$ & $109.03 \pm 5.35$ & $84.17 \pm 18.17$ & $2.56\text{E+}4 \pm 84.23$ & $371.65 \pm 10.81$ & $77.98 \pm 1.73$ \\ \hline
6 & $\mathbf{85.68 \pm 5.07}$ & $133.80 \pm 4.32$ & $123.60 \pm 5.25$ & $99.51 \pm 15.48$ & $6.71\text{E+}4 \pm 362.68$ & $1363.93 \pm 20.21$ & $93.84 \pm 1.26$ \\ \hline
7 & $\mathbf{100.24 \pm 5.96}$ & $120.23 \pm 5.06$ & $135.58 \pm 5.99$ & $118.91 \pm 15.67$ & $6.54\text{E+}4 \pm 214.22$ & $2317.88 \pm 15.54$ & $110.48 \pm 1.63$ \\ \hline

\end{tabular}
\end{table}

\newpage

\section{Additional Simulator Design Details}
\label{sec:simulator-design}

We present a queuing system testing framework. The main goals of the framework are:
1. Provide benchmarks for queuing algorithms
2. Easy to test and deploy with an OpenAI Gym Interface
3. Allow easy configuration of new custom queuing systems with a large degree of freedom

Our testing framework allows the following user interactions with intuitive interface
\begin{itemize}
    \item Defining a new queueing system or using one provided by our benchmark.
    \item Defining a policy that takes in observations and output queue priority prediction 
    \item Simulating queueing system trajectories with selected policies
\end{itemize}

We will detail each of these components of our framework below

\subsection{Define queuing system}
\label{sec:elements_of_queue}
\textbf{Ingredients of a queueing system} Our framework allows for flexible definition of queuing systems in a straightforward interface. A queuing system can be defined with the following descriptions:
\begin{itemize}
    \item \textbf{Network matrix}: a binary matrix that specifies which server can serve which queue
    \item \textbf{Network transition matrix}: what happens when a server finishes serving a job
    \item \textbf{Service rate matrix}: a matrix that specifies how fast a server can serve a queue. Time of service is drawn from a distribution specified by user using service rate matrix as parameter
    \item \textbf{Arrival rate of queues}: User can define arbitrary arrival pattern for queues as a Python function that inputs time and outputs time until next arrival. This feature allows simulation of time-varying arrivals. User can define arrival rate as a random distribution
    \item \textbf{Queue holding cost}: holding cost per unit of time for each job in each queue
    \item \textbf{Server pool}: We also allow user to define each class of server as a pool of server. When having many servers with the same characteristics, instead of creating many separate servers and inflating the size of network matrix, we allow users to specify a server pool number for each server class. This mechanism allows simulation of large-scale system without slowing the simulation.
\end{itemize}
Users can define these elements of a queuing sytems in a \texttt{.yaml} file and a \texttt{.py} file.

Here, we show an example \texttt{.yaml} file for configuring a criss-cross network:

\begin{verbatim}
name: 'criss_cross_bh'
lam_type: 'constant'
lam_params: {val: [0.9, .000001, 0.9]}
network: [[1,0,1],[0,1,0]]
mu: [[2,0,2],[0,1,0]]
h: [1,1,1]
init_queues: [0, 0, 0]
queue_event_options: [[1., 0, 0.],
                      [0., 0., 0.],
                      [0., 0, 1.],
                      [-1., 1., 0.],
                      [0., -1., 0.],
                      [0., 0., -1.],]
\end{verbatim}

\subsection{Defining a policy}
User can define a policy as a function that takes in queue length as observation and output a matrix that represents the policy's prediction of service priority. The matrix has the the same shape as network matrix (\# server $\times$ \#queues) that assigns priority to each server-queue pair. A policy can be either a static policy that decide priority based on observation with heuristics or contain a neural model to be trained. 

\subsection{Simulator design}
The simulator environment is structured as OpenAI Gym environment. We follow the design of the OpenAI Gym so that users can easily train and test a variety of reinforcement learning algorithms. Each simulation trajectory consists of a sequence of steps (defined in OpenAI Gym \texttt{step} format). For each simulation trajectory, the simulator maintain a number of information as states. 

\subsubsection{Simulator features}
We highlight some important features of our simulator below:
\begin{itemize}
    \item \textbf{Job-level tracking} The simulator tracks the states on the job level. We track service time for each job in a queue. At each step, we allocate to decide which job is being served on an individual job level. This mechanism makes it possible for multiple servers to serve a single queue and allows the modeling of parallel server systems.
    \item \textbf{Event-based simulation} Our simulator is event-based. Each step corresponds to one event: arrival in a queue or one job finishes being served. Prior works designed simulators with fixed time-interval for each step. In comparison, we can simulate trajectories with more uneven event intervals with higher speed by reducing wasting steps on intervals without any event. We also allow more precise time keeping.
    \item \textbf{Batch simulation} Our simulator allows simulation of multiple runs in parallel. Our parallelization implementation allow users to leverage accelerators like GPUs to accelerate simulations.
\end{itemize}

\subsubsection{States}
In each trajectory, the simulator keeps track a number of variables as states

Based on the elements of queue systems defined above, the simulator also has the capability of drawing a new service duration for a job and arrival duration for a queue. In addition, during a simulation run, the environment keeps track of
\begin{itemize}
    \item \textbf{Service time} Time until service finishes for a job
    \item \textbf{Arrival time} Time until next arrival occurs for a queue
    \item \textbf{Queue length} Length of each queue
\end{itemize}
At each step, the simulator updates service time and arrival time based on the event duration of the step. The simulator also has the capability to generate service time and arrival time for new jobs based on user specification of the queuing system. The simulator also updates queue lengths at each step based on the event occurred during the step.

\subsubsection{An event-based simulation step}
At each step, the simulator takes in action represented by the service priority matrix and returns the updated states in OpenAI Gym \texttt{step} function format. To simulate a step and obtain the output of the \texttt{step function}, our simulator decides an event that occurs based on the following procedure
\begin{itemize}
    \item \textbf{Converting service priority to action matrix} The step function takes in service priority prediction from the policy. The priority matrix can be a float matrix. The \texttt{step} function converts this priority matrix into an action matrix that specifies which servers should serve which queues. Users can customize how the assignment is done. Default implementation provides linear assignment, softmax, and Sinkhorn assignment.
    \item \textbf{Job-server allocation} The action matrix pairs server and queues. Our simulator then assigns each job in in a queue to servers that the action matrix decides to serve the queue through an \texttt{allocator} function. User can customized how the allocation is performed. The default allocator implementation selects the fastest serving servers and pair them with jobs with shortest service time remaining.
    \item \textbf{Select event} Based on the remaining service time for each job and remaining arrival time for each queue, the simulator decides the closest next event to be either (1) a job finishes being served or (2) a new job arrives for a queue.
    \item \textbf{Update states} Based on the event occurred, the simulator updates the states correspondingly. If a job finishes being served, the simulator removes the job from the queue. If the transition matrix specifies that the job in one queue goes to another queue after being served, an new job is created for the queue that the job transitions into. If a new arrival occurs, the simulator creates a new job for the queue and generate the new arrival time until next arrival in the queue. Finally, the simulator deducts the event duration from all service times and arrival times.
\end{itemize}

\section{Additional Experiment Details}

\subsection{Computational Resources}
\label{sec:comp-resources}
We run all our experments on an AMD EPYC 7513 32-Core Processor.

\end{document}